\documentclass{article}

\usepackage{amsthm}

\usepackage{float} 
\usepackage{multirow}
\usepackage{booktabs}
\usepackage{xcolor}
\usepackage{amsmath, amssymb}
\usepackage{threeparttable}
\usepackage{tabularx}
\usepackage{subcaption}
\usepackage[preprint]{neurips_2026}

\usepackage[utf8]{inputenc} 
\usepackage[T1]{fontenc}    
\usepackage{hyperref}       
\usepackage{url}            
\usepackage{booktabs}       
\usepackage{amsfonts}       
\usepackage{nicefrac}       
\usepackage{microtype}      
\usepackage{xcolor}         
\usepackage{graphicx}

\title{Not All Starting Points Are Equal: Pre-trained Priors and Their Outsized Impact on Person Identification}

\author{%
  Thomas M Metz \\
  School of Behavioral and Brain Sciences\\
  The University of Texas at Dallas\\
  Richardson, Texas, USA \\
  \texttt{thomas.metz@utdallas.edu} \\
  \And
  Matthew Q Hill \\
  School of Behavioral and Brain Sciences\\
  The University of Texas at Dallas\\
  Richardson, Texas, USA \\
  \texttt{mqh100020@utdallas.edu,} \\
  \And
  Alice J O’Toole \\
  School of Behavioral and Brain Sciences\\
  The University of Texas at Dallas\\
  Richardson, Texas, USA \\
  \texttt{otoole@utdallas.edu} \\
  \And
}

\begin{document}

\maketitle

\begin{abstract}
Recent years have seen an explosion of diverse general purpose pre-training methodologies for computer vision. However, the impact that these pre-training methodologies have on person identification tasks (re-id) remains under-explored. We show that under equated domain adaptation pipelines, there is dramatic variance in person identification outcomes using different starting models (architectures and pre-trained weights). We show that a range of intuitive explanations for differing downstream performance on a range of re-id tests are insufficient and propose that pre-trained weights serve as a strong prior to the weights learned during domain adaptation. This framework allows for domain adapted solutions to be viewed as a maximum probability point estimate of the Gibbs posterior with the pre-trained weights acting as a prior. Under this framework, we show that large, pre-trained foundation models with simple domain adaptation achieve SOTA solutions on a range of re-id datasets (Market, PRCC, DeepChange, BTS) with solutions that are very close in the parameter space to the starting parameters. Moreover, we perform ablations on these solutions and show that they can be reached with small transfer sets and with varying transfer datasets but are sensitive to choice of optimizer, weight-decay, and loss function. Ultimately, we propose that the  simple approach of direct fine-tuning using large vision foundation models (CLIP, Dino, EVA, AIM, etc.) needs to serve as an important baseline for future work in re-id. 
\end{abstract}

\section{Introduction}

Whole-body person identification 
has been approached in two ways.
Short-term person re-identification (re-id) tracks a person
in a closed environment (e.g., train station). In this case, transient appearance cues, like clothing, can support effective re-id.
In long-term person re-id,
the goal is
to identify people
over multiple time points and across changing environments. 
The subject may change their appearance (e.g., clothes) and there may be substantial
differences in imaging conditions (e.g., distance, view).
Long-term person re-id requires  the encoding of person-based attributes that are independent of short-term  cues (e.g., clothes-change re-id, cc re-id). 
Although short-term re-id has been  studied (for a review \cite{9336268}),
work on long-term re-id has emerged only recently with the increased availability of large-scale clothing-change datasets (e.g., \cite{yu2020cocas, cornett2023expanding, jager2025expandingbriardatasetcomprehensive}). Critically, successful algorithms for short-term re-id often do not transfer to the clothes-change scenario. Market-1501\cite{7410490} can be classified as short-term re-id due to its clothing constancy. PRCC \cite{9710001} falls between short and long term re-id because clothing is not a reliable cue but environment and imaging conditions are controlled. BTS \cite{cornett2023expanding} and DeepChange \cite{xu2022deepchangelargelongtermperson} qualify as long-term re-id, unconstrained re-id where identification must take place across varying clothing sets, diverse environments, and extreme imaging conditions.

Sources of identifying information for long-term re-id include the face, body, and gait \cite{hahn2016dissecting}. We focus on performing identification from whole-body still images. Identification is performed with whole, partial, or occluded body images, where   clothing is not a reliable identity cue and pose is not controlled. This work fills an important gap in current understanding of person identification by showing that pre-training plays a critical, currently understated role in all re-id varieties.  

\section{Related Works}
Existing work in re-id almost always relies on pre-trained models. There are two methods to incorporate pretraining into identification pipelines. The first  applies domain transfer learning to object-pretrained networks (often ImageNet \cite{russakovsky2015imagenet}). The second  uses a foundation model to provide a supervisory signal, and commonly avoids object pretraining, because short term re-id work suggests that self-supervised pretraining is superior \cite{luo2021selfsupervisedpretrainingtransformerbasedperson}. Despite relying heavily on pre-trained models, the extent to which the starting pretrained model impacts outcomes
is unknown.

\textbf{Object Transfer Approaches:}\label{non-foundation}
Early studies on cc re-id utilized CNNs pretrained on ImageNet and focused directly on extracting clothing-agnostic features. Adding a clothes shape distillation module to a pretrained ResNet \cite{he2015deepresiduallearningimage} proved effective on the Long Term Clothes Change (LTCC) dataset \cite{qian2020longtermclothchangingpersonreidentification}. In a similar vein, a set of parallel stream 
ImageNet-pretrained CNNs with clothing status awareness improved SOTA results on the Person Re-identification with Clothing Change (PRCC) dataset \cite{9710001}. A third parallel stream architecture  applied causality-inspired clothes debiasing to ImageNet-pretrained CNNs and improved SOTA results on both PRCC and LTCC \cite{10203842}. In Clothes-based Adversarial Loss (CAL), adversarial training was used to force the backbone to rely on clothing-irrelevant features. CAL improved SOTA results on most common clothes-change datasets \cite{gu2022clothes}. Additional work in extracting clothing agnostic features introduced Clothing-Change Feature Augmentation (CCFA) to remedy the limited availability of extensively variable clothes-change data for the same ID. This technique yielded a notable performance boost on LTCC \cite{han2023clothing}. 

Work on unconstrained, long term re-id \footnote{A variant of cc re-id that uses extreme imaging conditions like photos taken from 1000m at any pitch and yaw.} has primarily  used data-driven approaches with the goal of extracting clothing-agnostic features---typically incorporating ImageNet pretraining. The Non-linguistic Core ResNet Identity Model (NLCRIM) \cite{myers2023recognizing} applied domain transfer learning with clothes-change data to an ImageNet-pretrained ResNet101 \cite{he2015deepresiduallearningimage} with no specialized clothing modules or loss functions. This  performed best on images taken
at altitude and long range 
in the Biometric Recognition and Identification at Altitude and Range (BRIAR) dataset \cite{cornett2023expanding, jager2025expandingbriardatasetcomprehensive}. The Linguistic Core ResNet Identity Model (LCRIM) \cite{myers2023recognizing} was similar, but added intermediate training to predict human body descriptors from images. LCRIM performed best on close-range data. 
Fusing  NLCRIM and LCRIM outputs increased accuracy in nearly all cases.
BRIAR-Net adapted a ResNet50 \cite{he2015deepresiduallearningimage} and relied on data and loss functions (triplet-loss + cross entropy loss) to learn clothing-agnostic features \cite{huang2024whole}. BRIAR-Net surpassed previous methods on the BRIAR data.

Modern object transfer approaches to cc re-id now use pretrained Vision Transformers \cite{dosovitskiy2021imageworth16x16words}, due to their ability to model long-range dependencies  \cite{He_2021_ICCV, luo2021selfsupervisedpretrainingtransformerbasedperson, sharma2021personreidentificationlocallyaware, 10.1145/3474085.3475202}. 
As more clothes-change datasets have 
emerged\footnote{Note: The BRIAR dataset has expanded continuously since its initial publication, enabling more large-scale unconstrained re-id work.}, ViTs quickly became the preferred starting point 
for cc re-id. Ad-ViT proposed a cc re-id strategy based on an ImageNet pretrained ViT and exploited descriptors that are invariant to clothing as training guidance \cite{9959509}. This model yielded SOTA results on LTCC and on the Non-overlapping Knowledge-aware dataset for Unlimited person re-identification under Persistent clothing changes 
(NKUP) \cite{NKUP}. Other work with ViTs proposed hybrid models, where a ViT + body shape motion feature framework achieved SOTA results on PRCC and LTCC \cite{9707584}. 

The Body Identification from Diverse
Datasets (BIDDS) model \cite{myers2025unconstrainedbodyrecognitionaltitude} proposed a multi-stage training strategy on 1.9 million body images---most with clothing change. BIDDS first adapted an ImageNet pretrained ViT to person re-id. Next, the model was fine-tuned with  unconstrained cc re-id data. BIDDS performed well on a broad range of traditional, cc, and unconstrained cc re-id tasks, but was surpassed by Swin-BIDDS \cite{myers2025unconstrainedbodyrecognitionaltitude}  (a comparable training strategy utilizing the Swin backbone \cite{liu2021Swintransformerhierarchicalvision} and a larger image size in training).
BIDDS and Swin-BIDDS both surpassed similarly-trained CNNs \cite{myers2025unconstrainedbodyrecognitionaltitude, metz2025dissectinghumanbodyrepresentations}. Both models achieved near SOTA results on cc re-id via a simple, direct fine-tuning of an ImageNet-1k pre-trained ViT and an ImageNet-1k Swin-ViT with unconstrained whole-person data.

\textbf{Foundation-Supervised Approaches:} \label{Foundation}
The broad success of vision foundation models has led to their use in cc re-id. Because foundation models can generate stable and generalizable feature spaces, they have become a popular tool to provide supervisory signals to student networks.

{\it Contrastive Language-Image Pretraining (CLIP)}.
Linguistically-guided approaches to person re-id have used CLIP \cite{radford2021learningtransferablevisualmodels} ViTs. In one CLIP-based model,  a two-stage training approach  exploited an index label as linguistic guidance and incorporated visual information \cite{li2023clipreidexploitingvisionlanguagemodel}.
As noted in that work, direct fine-tuning of a CLIP vision encoder performs well in short-term re-id problems. 
Later work in clothes-change re-id attempted to remedy the CLIP image encoder's over-reliance on clothing information by utilizing custom modules to guide the extraction of clothing-agnostic information \cite{li2024clipdrivenclothagnosticfeaturelearning}. Other research attempted to generate better linguistic guidance via synthetic descriptors \cite{han2024clipscgisynthesizedcaptionguidedinversion} and to construct a framework to generate domain-invariant and domain-specific linguistic prompts \cite{zhao2025cilpfgdiexploitingvisionlanguagemodel}. Notably, CLIP3DReID used CLIP itself to generate labels by exploiting contrasting clothing-invariant descriptors. 
It used optimal transport theory to align a student model’s
local visual features with shape-aware tokens derived from
CLIP’s linguistic output. Additionally, it aligned a student
model’s global visual features with features from the CLIP
image encoder and the 3D SMPL identity space \cite{loper2023smpl}. This approach showed strong results on long-term re-id \cite{10655020}. 



{\it \textbf{E}xplore the limits of \textbf{V}isual representation at sc\textbf{A}le-02 (EVA-02)}.
This model \cite{Fang_2024}  combines pretraining innovations from Natural Language Processing (NLP) and yields SOTA performance on multiple image tasks
(e.g., ImageNet classification, object detection, and semantic segmentation). 
A descendant of a previous network \cite{fang2023eva}, 
EVA-02 uses the TransformVision (TrV) architecture, which modifies a traditional ViT encoder by adding a SwiGLU Feed Forward Network \cite{ramachandran2017searchingactivationfunctions, dauphin2017languagemodelinggatedconvolutional, shazeer2020gluvariantsimprovetransformer, hendrycks2023gaussianerrorlinearunits}; sub-layer normalization \cite{wang2022foundationtransformers}; rotary positional embeddings (RoPE) \cite{su2023roformerenhancedtransformerrotary}; and xavier normal weight initialization \cite{pmlr-v9-glorot10a}. EVA-02 is pretrained with Masked Image Modeling (MIM) using a CLIP teacher. Notably, EVA-02 is trained with a larger CLIP teacher than has been used previously with MIM.

EVA-02 has been utilized for cc re-id in the Masked Attribute Description Embedding (MADE) framework \cite{peng2024maskedattributedescriptionembedding}. MADE proposes a framework that integrates visual appearance and attribute descriptions, building on the transform vision backbone \cite{Fang_2024}. Clothing change is dealt with by utilizing a Description Extraction and Mask module. MADE masks clothing features and  performs well on PRCC, LTCC, Celeb-reID-Light \cite{huang2019celebrities}, and LaST \cite{shu2021large}. 

{\it Additional Foundations}
Other popular foundation models like DINOv2\cite{oquab2023dinov2} and AIMv2\cite{fini2024multimodalautoregressivepretraininglarge} do not appear directly in the re-id literature. 

\textbf{Impact of pre-training on re-id:}
Despite a wide range of work either directly domain adapting or using pre-trained models to supervise learning, to our knowledge, no work directly addresses the role that pre-training plays in resulting solution quality. It is not intuitive that a model's starting point should have a large impact on its downstream performance in person identification tasks. Identification tasks are fundamentally different from classification tasks. Even neuroscience suggests that specialized areas perform recognition (Fusiform Face Area, Extrastriate Body Area). However, we believe this to be a critical gap to understanding re-id problems because if pre-training plays a large role, benchmarking across re-id methodologies becomes challenging.

\begin{figure}
    \centering
    \includegraphics[scale=0.33]{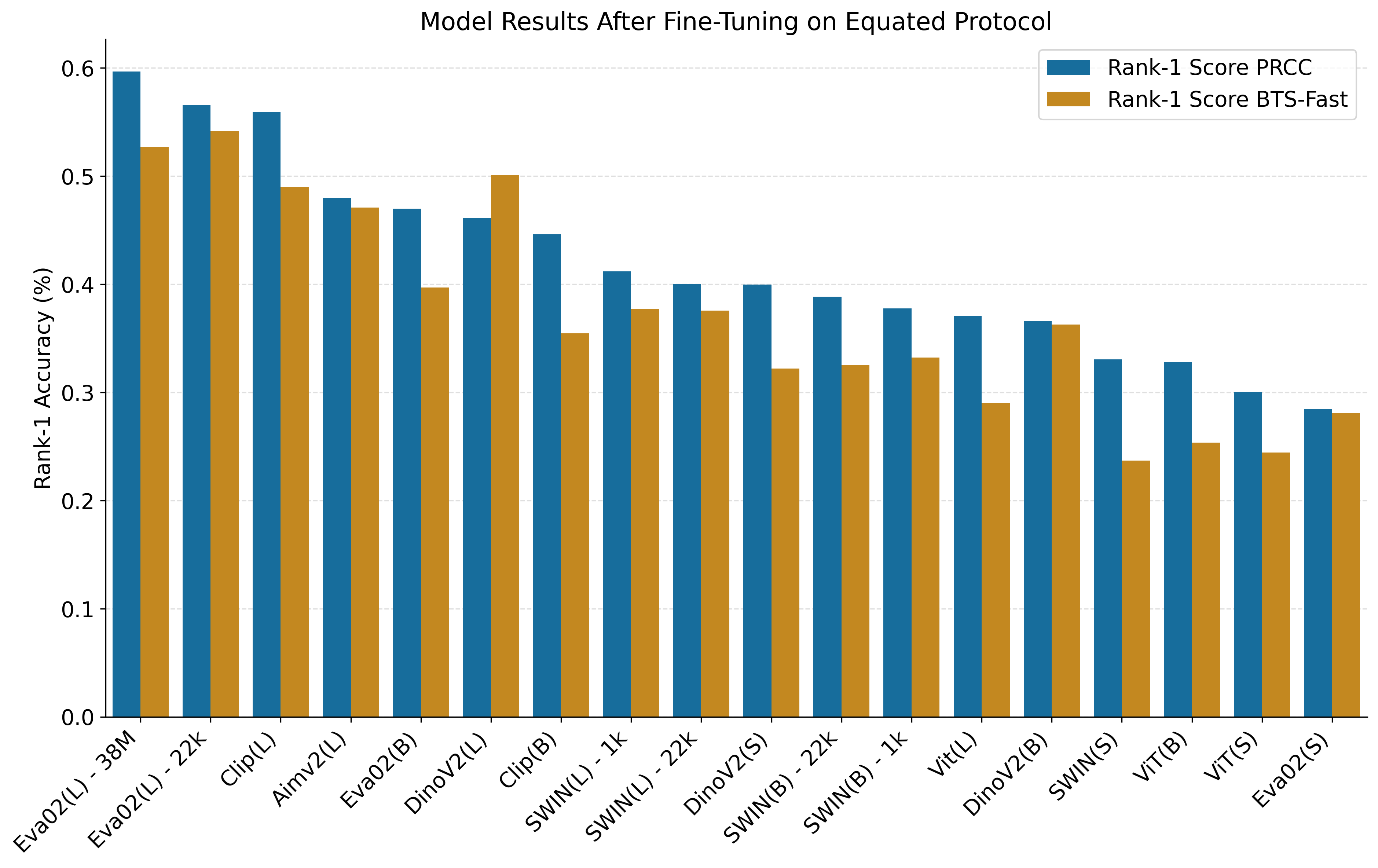}
    \caption{Choosing different starting models has a strong impact on re-id outcomes using an equated transfer learning protocol, even when comparing identical architectures.}
    \label{all_model_results_prcc}
\end{figure}

\section{Methods}
To address this gap, we perform domain adaptation on 18 pre-trained models that are common in the person identification  literature. We use a very simple transfer protocol that we call:  the ``study'' protocol. The ``study'' protocol is designed to find solutions close to the pre-trained model's starting parameters $\theta_0$. It uses an online triplet loss function with batch hardest mining, which is a common and effective choice for all re-id problem types. It uses a margin of $.35$; a small batch size (40) to promote generalization; and 4 positives are sampled per id. A small learning rate is used ($7.5\times10^{-6}$) to search for solutions near $\theta_0$ and small weight decay ($1\times10^{-6}$) to discourage overfitting while not strongly penalizing large starting parameters. It uses an ADAM optimizer and  random horizontal flip, color jitter, random grayscale, and gaussian blur transformations. This method is intentionally simple and designed to find solutions near pre-trained parameters, $\theta_0$; it is a means to our contribution, not the contribution itself.

All training and experiments are conducted on two Nvidia-A6000s with 48Gb VRAM; system RAM is 512 GB. An AMD Ryzen Threadripper PRO 3945WX CPU is used (12 cores, 24 threads, up to 4.43 GHz).

\begin{figure}
    \centering
    \begin{tabular}{c}
    \includegraphics[scale=0.35]{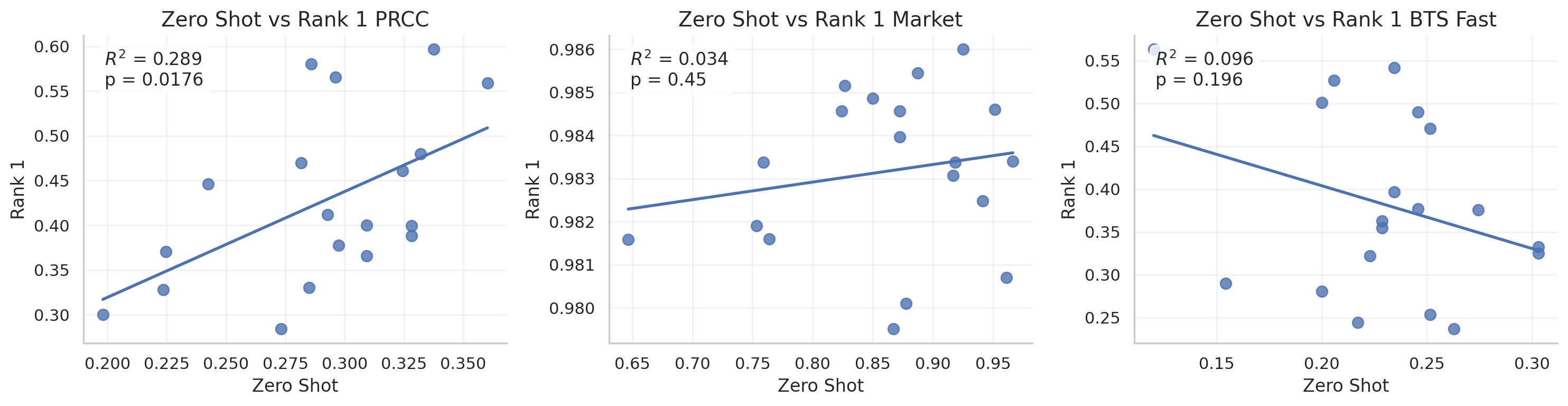} \\
    \includegraphics[scale=0.35]{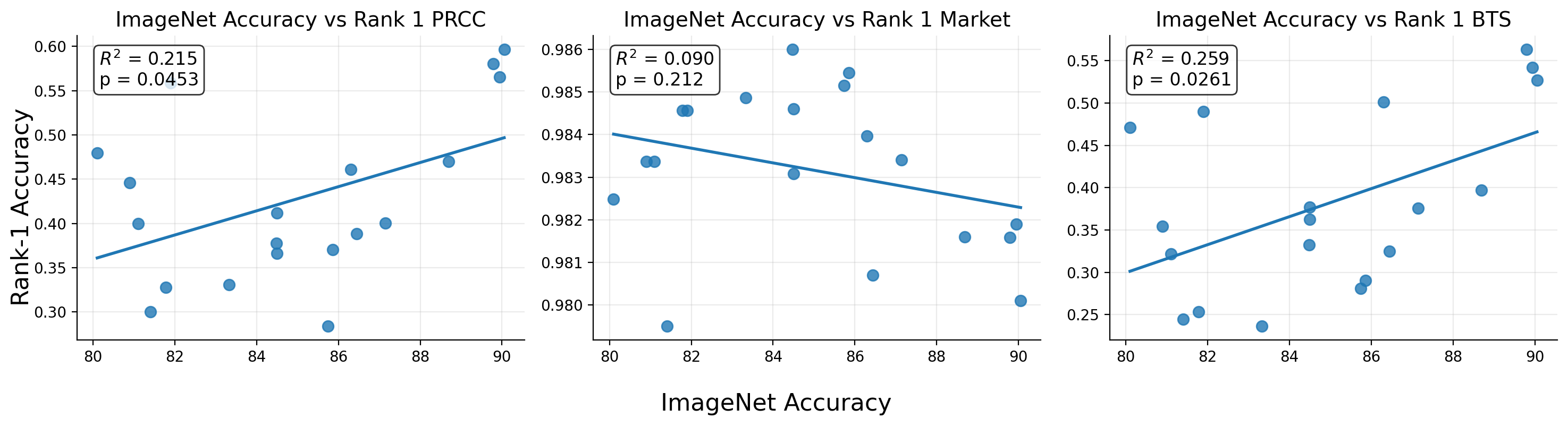} \\
    \end{tabular}
    \caption{(top row) Zeroshot Performance is a weak predictor of re-id performance after domain adaptation. (bottom row) ImageNet Performance is a weak predictor of re-id performance after domain adaptation.}
    \label{ZSIN_VS_R1}
\end{figure}

\textbf{Data:}
The study protocol performs a one stage transfer learning with a given dataset. Study-pooled (abbreviated S-Pool) uses pooled data from the BRIAR Research Set (BRS) (859k images of 1.2k identities) \cite{cornett2023expanding, jager2025expandingbriardatasetcomprehensive}, Kitware BRC (KWBRC) (30.k images of 149 identities)\cite{KWBRC}, and DeepChange (75k images of 450 identities)\cite{xu2022deepchangelargelongtermperson}. Study PRCC (S-PRCC) uses PRCC training data. Study Market-1501 (S-Market) uses Market-1501 training data. The one stage transfer seeks to eliminate confounding variables introduced by more performant multi-stage transfer approaches.

\section{Results}
\textbf{Different pre-trained models perform vastly different after domain adaptation}

We perform domain adaptation via the study-pooled protocol on 18 commonly used pre-trained models with varying size, architecture, and starting weights. Results shown in Figure \ref{all_model_results_prcc} demonstrate that the rank 1 accuracy on PRCC and BRIAR Test Set (BTS) data \footnote{BTS-Fast references a subset of 372 identities and 14,880 images. Probe and gallery sets are strictly in different clothing sets.}  varies across starting model. Very strong models ($\approx.6$ PRCC rank 1 without fine-tuning on PRCC) and very mediocre models ($\approx.3$ PRCC rank 1 without fine-tuning on PRCC) result. We conclude that under equated training protocols, different pre-trained models yield vastly different re-id outcomes after domain transfer.

\textbf{Intuitive explanations fail:}
We tested five intuitive explanations as to why different pre-trained models yield different outcomes and conclude that all are insufficient. All of the following analyses are of pre-trained models transferred with ``S-Pool'' that are shown in Figure \ref{all_model_results_prcc}.

{\it 1. Architecture does not explain the difference in outcomes.}
CLIP(L) and ViT(L) share a ViT-large backbone with different pre-training. CLIP(L) substantially outperforms ViT(L), with a 20-point boost in rank 1 performance on BTS data and a 19-point boost on PRCC data. Likewise, DINOv2(L) substantially outperforms ViT(L) though both use a ViT-large backbone; CLIP(B) substantially outperforms ViT(B) despite both using a ViT-base backbone; and DINOv2(S) substantially outperforms ViT(S) despite both using a ViT-small backbone. 

{\it 2. Model size does not explain the difference in outcomes:}
CLIP(B) outperforms ViT(L) by 6 and 8 rank-1 points on the BTS and PRCC datasets, respectively. Similarly, Eva02(B) surpasses both Swin(L) and ViT(L), while DinoV2(S) outperforms Swin(B), ViT(L), and ViT(B) on PRCC, and ViT(L) and ViT(B) on BTS. Small models can outperform larger ones after transferring.

{\it 3. ImageNet performance does not explain the difference in outcomes:}
We regress each model's ImageNet accuracy vs.\ its rank-1 accuracy on PRCC, Market-1501, and BTS data after domain adaptation under the equated ``S-pool''. Across each re-id dataset there is little relationship between a model's ImageNet performance and its performance after domain adaptation (See Figure \ref{ZSIN_VS_R1} bottom row). Stronger object models, at least measured by proxy of ImageNet performance, do not necessarily transfer to produce stronger re-id models. 

{\it 4. ZeroShot performance does not explain the variance in outcomes:}
We regress each model's zeroshot rank-1 accuracy vs. its rank-1 accuracy on PRCC, Market-1501, and BTS data after domain adaptation. Across each re-id dataset, there is little relationship between zeroshot performance and performance after domain adaptation. We conclude that models that are weak in zeroshot mode are not necessarily weak after transferring.

{\it 5. Initial model features are insufficient to perform challenging re-id tasks:}
Across all models, zero-shot features with a trained classifier perform substantially worse than each model after domain adaptation. We trained two classifiers on top of frozen model features. First, we trained a linear classifier using the identical approach established in the study-pooled protocol with the exception that we kept all model features (except for a randomly initialized linear head) frozen. Second, we used the principal component ablation + oracle algorithm proposed in \cite{metz2025dissectinghumanbodyrepresentations}.  This algorithm has been shown to lead to boosts in re-id performance with frozen features. Regardless of the classifier added on top of frozen features, rank-1 performance on BTS data is more than $50\%$ worse than performing a full fine-tuning. In the supplementary materials, we  show likewise that the trained classifiers perform substantially worse than the domain adapted models on PRCC data. Therefore, semantically rich zero-shot features alone cannot explain models' re-id performance after domain adaptation. 



\begin{figure}
    \centering
    \includegraphics[scale=0.4]{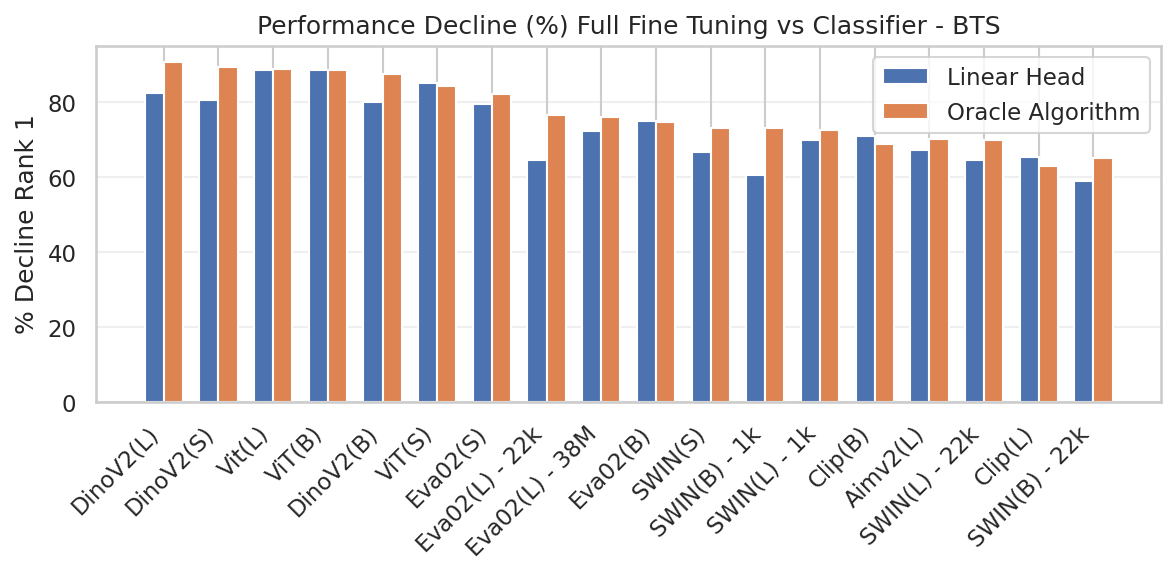}
    \caption{Frozen features from pre-trained models are insufficient to perform re-id problems even with classifiers trained on top of them.}
    \label{fig:total_decline}
\end{figure}



\begin{table}[htbp]
\scriptsize
\centering
\begin{tabular}{lllccc ccc}
\toprule
Model & Pre-training & Pre-training data
& \multicolumn{3}{c}{Study Pooled}
& \multicolumn{3}{c}{Study PRCC} \\
\cmidrule(lr){4-6} \cmidrule(lr){7-9}
& & 
& $L_2$ & $\sigma$ & loss
& $L_2$ & $\sigma$ & loss \\
\midrule
Swin(S) & Supervised & IN22k, IN1k    & 0.0038 & 0.02 & 0.422 & 0.0038 & .01 & 0.028\\
ViT(S) & Supervised & IN21k           & 0.0065 & 0.05 & 0.065 & .0063 & .01 & 0.019 \\
ViT(B) & Supervised & IN21k           & 0.0039 & 0.03 & 0.767 & 0.0023& .00& 0.005\\
ViT(L) & Supervised & IN21k           & 0.0013 & 0.03 & 0.609 & 0.0008 & .00 & 0.009\\
Swin(B) & Supervised & IN1k           & 0.0062 & 0.09 & 1.811 &0.0024 &.01 & 0.011\\
Swin(L) & Supervised & IN1k           & 0.0036& 0.05 & 0.087 & 0.0015 & .00 & 0.012 \\
Swin(B) & Supervised & IN22k          & 0.0032 & 0.03 & 1.435 & 0.0024 & .01 & 0.009\\
Swin(L) & Supervised & IN22k          & 0.0033 & 0.04 & 0.457 & 0.0015 & .00 & 0.008\\
DinoV2(S) & DINO & LVD-142M           & 0.0049 & 0.01 & 0.111 & 0.0062 & .00 & 0.010\\
DinoV2(B) & DINO & LVD-142M           & 0.0020 & 0.02 & 0.037 &0.0022& .00 & .007\\
DinoV2(L) & DINO & LVD-142M           & 0.0010 & 0.04 & 0.026 & 0.0007 & .00& .003\\
CLIP(B) & CLIP & CLIP-WIT             & 0.0018 & 0.02 & 0.082 & 0.0021 & .00 & .002\\
CLIP(L) & CLIP & CLIP-WIT             & 0.0007 & 0.01 & 0.030 &0.0007 &.00 & 0.013\\
Eva02(S) & MiM + FT & IN22k, IN1k     & 0.0062 & 0.07 & 0.335 & 0.0061 & .01 & 0.005\\
Eva02(B) & MiM + FT& IN22k, IN1k      & 0.0019 & 0.01 & 0.122 &0.0021 & .00 & 0.007\\
Eva02(L) & MiM + FT& IN22k, IN1k      & 0.0012 & 0.03 & 0.033 & 0.0007 & .00 & 0.001 \\
Eva02(L) & MiM + FT & 38m, IN21k, IN1k& 0.0003 &0.01 & 0.027 & 0.0007 & .00 &0.001\\
AIMV2(L) & AIM & AIM 12B              &0.0008 & 0.01& 0.047 &0.0007& .00& 0.018\\
\hline

\end{tabular}
\caption{Average parameter distance before and after training. Most models exhibit low loss solutions after transferring where parameters move on average a few hundredths of their original standard deviation. The shown loss is estimated from 30 training batches using Triplet loss with batch hardest mining and a margin of .35.}
\label{tab:param_distance}
\end{table}

\textbf{Pre-training as a prior:}
The study protocol is designed to find person re-id solutions near starting pre-trained parameters $\theta_0$. In Table \ref{tab:param_distance} shows that in absolute terms, and standardized with regards to parameter scale, the learned solutions are very close to $\theta_0$ for two different transfer datasets. Thus, the pre-trained parameters act as a strong prior under the study protocol. 

{\it Low bias priors.}
Given that $\theta_0$ acts as a strong prior, the reported loss data in Table \ref{tab:param_distance} shows that some priors exist near low loss solutions. For example, models like CLIP (L), Vit(S), SWIN(L), etc. achieve very low loss on the pooled training set while their parameters move minimally. Moreover, all models achieve a low loss on PRCC training data with very little parameter movement. Thus, using a triplet loss objective, some pre-trained models offer priors that exist very close to low loss solutions on the most challenging re-id problems (like BRS and DeepChange), and many pre-trained models offer such priors for less challenging---but still clothing change---re-id tasks (like PRCC).

{\it Low variance priors:}
Given that some low performing models (ViT(S), SWIN(L)) achieve very low training loss on the pooled dataset and many low performing models (ViT(S), Eva02(S), ViT(B), etc.) achieve very low loss on the PRCC dataset, the role of a pre-trained prior is not limited to encouraging low bias solutions. Rather, some priors lead to solutions that achieve very low training loss and generalize  well. These  are primarily reached by vision foundation models (e.g., CLIP(L), DinoV2(L), Eva02(L)). Intuitively, since vision foundation models learn initial features that can perform a broad range of tasks,  new parameters very close in the feature space may generalize well. 

{\it Re-id adapted foundations are point estimates for the Gibbs posterior:}
The Gibbs posterior extends the standard posterior distribution by allowing the substitution of the likelihood function with an empirical risk function \cite{Bissiri_2016}. This enables a Bayesian-style framework for understanding fine-tuning without specifying a statistical model.

Assume a network is randomly initialized with parameters
{\small
\begin{equation}
\theta \sim N(0_p, \Sigma_p),
\end{equation}
}
and for some loss function $L^{(1)}$ and data $D^{(1)}$, there exists
{\small
\begin{equation}
\hat{\theta}^{(1)} = \arg\min_{\theta} L^{(1)}(\theta, D^{(1)}).
\end{equation}
}
Here, $\hat{\theta}^{(1)}$ serves as the point estimator for the pre-training objective.

Now consider a new loss $L^{(2)}$ and new data $D^{(2)}$, and define the empirical risk as
{\small
\begin{equation}
R(\theta) = \frac{1}{n} \sum L^{(2)}_\theta(x_i).
\end{equation}
}

The Gibbs posterior is given by
{\small
\begin{equation}
\pi(\theta_1 \mid D^{(2)}) \propto \exp(-\omega \cdot n \cdot R(\theta)) \cdot \pi(\theta_1).
\end{equation}
}

Conditioning on $\theta_0 = \hat{\theta}^{(1)}$, we have
{\small
\begin{equation}
\pi(\theta_1 \mid D^{(2)}, \theta_0) \propto \exp(-\omega \cdot n \cdot R^{(1)}(\theta)) \cdot \pi(\theta_1 \mid \theta_0).
\end{equation}
}

Assuming
{\small
\begin{equation}
\pi(\theta_1 \mid \theta_0) \sim N(\theta_0, \Sigma),
\end{equation}
}
this becomes
{\small
\begin{equation}
\pi(\theta_1 \mid D^{(2)}, \theta_0) \propto \exp(-\omega \cdot n \cdot R^{(1)}(\theta)) \cdot \exp\left[-\frac{1}{2} (\theta_1 - \theta_0)^T \Sigma^{-1} (\theta_1 - \theta_0)\right].
\end{equation}
}

Maximizing the log is then equivalent to minimizing
{\small 
\begin{equation}
\min \left[ w L_{\theta}^{(2)}(X) + \frac{1}{2} (\theta_1 - \theta_0)^T \Sigma^{-1} (\theta_1 - \theta_0) \right].
\end{equation}
}

In other words, the initialization $\theta_0$ acts as a regularizing term on the solution to the downstream loss function.

Most machine learning applications directly minimize $R$ only. However, as empirically shown in Table \ref{tab:param_distance}, when performing domain adaptation to re-identification, foundation models can achieve near-zero training risk while also remaining very close to the starting parameters. In this regard, foundation-adapted solutions to re-id problems, at least under a triplet loss objective, can be viewed as an approximate estimate of the solution obtained by maximizing the probability of the Gibbs posterior.

\textbf{Foundation models achieve SOTA results under domain adaptation}

To contextualize the impact of pre-training, we take foundation models and extend the study protocol by adding a second fine-tuning step on two common re-id datasets (PRCC and Market-1501) for comparison to re-id benchmarks. Foundation models achieve SOTA level solutions under this direct fine-tuning. Notably, CLIP(L) achieves 98.43 Rank-1 on Market-1501, and 73.30 Rank-1, 70.20 mAP on PRCC. On the very challenging unconstrained BTS 5 test set, foundation models fair even better, improving SOTA by over 14 points on rank 1 and over 15 points on TAR@FAR $10^{-3}$. The summed cosine fusion of adapted foundation models improves SOTA even further, outperforming any single model on all tested metrics for PRCC, DeepChange, and BTS 5. We conclude that the role of the pre-trained prior is dramatically understated for re-id problems, because, with simple domain adaptation, the right prior achieves new SOTA.

\begin{table*}[ht]
\begin{threeparttable}[b]
\centering
\caption{Fine-tuning existing general vision foundation models surpasses existing benchmarks on nearly all metrics for all datasets. The performance gain is nearly 5 points in Rank 1 on PRCC, over 40 points in Rank 1 on DeepChange, and 23 points in mAP on DeepChange. Best individual model in bold for each column; Fusion scores that exceed the best individual model scores are also in bold.}
\scriptsize
\label{tab:model_comparison}
\begin{tabularx}{\textwidth}{l l *{6}{>{\centering\arraybackslash}X}}
\toprule
\textbf{Model} & \textbf{Venue} & 
\multicolumn{2}{c}{\textbf{Market-1501}} & 
\multicolumn{2}{c}{\textbf{PRCC}} & 
\multicolumn{2}{c}{\textbf{DeepChange}} \\
\cmidrule(lr){3-4} \cmidrule(lr){5-6} \cmidrule(lr){7-8}
& & R1 & mAP & R1 & mAP & R1 & mAP \\
\midrule
Eva02-L & - & 98.22 & 88.14 & 72.31 & 68.90 & $\textbf{97.52}$ & 
\textbf{42.78} \\

CLIP-L & - & \textbf{98.43} & 84.33 & \textbf{73.30} & \textbf{70.20} &96.85 & 37.54 \\

DINOv2-L & - & 98.25 & 85.31 & 70.99 & 66.99 & 97.23 & 39.55 \\

AIMv2-L & - & \textbf{98.43} & 83.71 & 64.35 & 63.09 & 96.67 & 36.78 \\

\hline\hline
        Foundation Fusion & - & 98.40 & 89.87 & \textbf{78.30} & \textbf{74.77} & \textbf{98.14} & \textbf{45.85} \\
 \hline\hline
VILLS \cite{huang2025vills} & WACV 25  & 96.8 & \textbf{92.9} & 58.4 & 55.0 & - & - \\
DIFFER \cite{Liang_2025_CVPR} & CVPR 25  & - & - & 68.5 & 64.7 & - & - \\
 SOLIDER† \cite{chen2023beyond} & IEEE/CVF 23  & 96.9 & 93.9 & 50.1 & 49.9 & - & - \\
PASS \cite{10.1007/978-3-031-19781-9_12} & ECCV 22 &  96.8 &  92.3 & 52.4 & 53.4 & - & - \\
CCFA \cite{han2023clothing} & CVPR 23 &  - &  - & 61.2 & 58.4 & - & - \\
BSGA+CRE \cite{Mu_2022_BMVC} & BMVC 22 &  - &  - & 61.8 & 58.7 & - & - \\
MADE \cite{10814721} & IEEETM  &  - &  - & 64.3 & 59.1 & - & - \\
AIM \cite{10203842} & CVPR 23 &  - &  - & 57.9 & 58.3 & - & - \\
CAL \cite{gu2022clotheschangingpersonreidentificationrgb} & CVPR 22 & 94.7 & 87.5 & 55.2 & 55.8 & 54.0 & 19.0 \\

\bottomrule
\end{tabularx}
\end{threeparttable}
\end{table*}

\begin{table*}[ht]
\begin{threeparttable}[b]
\centering
\caption{Model Performance on the Briar Test Set. Best performance for the individual models is in boldface. Fusion is superior to all individual models and is also in bold. T@F refers to True Accept Rate @ False Accept Rate}
\scriptsize
\label{tab:model_comparison_briar}
\begin{tabularx}{\textwidth}{l l l  *{4}{>{\centering\arraybackslash}X}}
\toprule
\textbf{Model} & \textbf{Venue} & \multicolumn{4}{c}{\textbf{BTS 5}} \\
\cmidrule(lr){3-6}
 &  & \textbf{R1} & \textbf{R20} & \textbf{T@F $10^{-4}$} & \textbf{T@F $10^{-3}$} \\
\midrule

    Eva02-L & - & \textbf{67.9} & \textbf{93.2} & 31.5 & \textbf{59.8} \\

AIMv2-L & - & 61.0 & 90.2 & 27.6 & 49.82 \\

CLIP-L & - & 67.0 & 92.4 &  \textbf{36.1} & 57.8\\

DINOv2-L & - & 61.0 & 90.4 & 29.4 & 52.7 \\
\hline\hline
Foundation Fusion & - & \textbf{69.5} & \textbf{93.9} & \textbf{39.1} & \textbf{63.7} \\

\hline\hline
BIDDS \cite{myers2025unconstrainedbodyrecognitionaltitude} & FG 2025 & 45.1 & 85.7 & 18.7 & 38.3 \\
Swin-BIDDS \cite{myers2025unconstrainedbodyrecognitionaltitude} & FG 2025 & 50.6 & 88.9 & 21.8 & 44.6 \\
VILLS \cite{huang2025vills} & WACV 25 & 53.1 & 91.8 & 21.7 & 43.6\\
\bottomrule
\end{tabularx}
\end{threeparttable}
\end{table*}

\section{Analysis}
Given that foundation models provide priors that can achieve SOTA solutions on a range of re-id tests, we ablate components of our simple transfer protocol to learn more about these solutions. 

\begin{figure}
    \centering
    \includegraphics[scale=0.4]{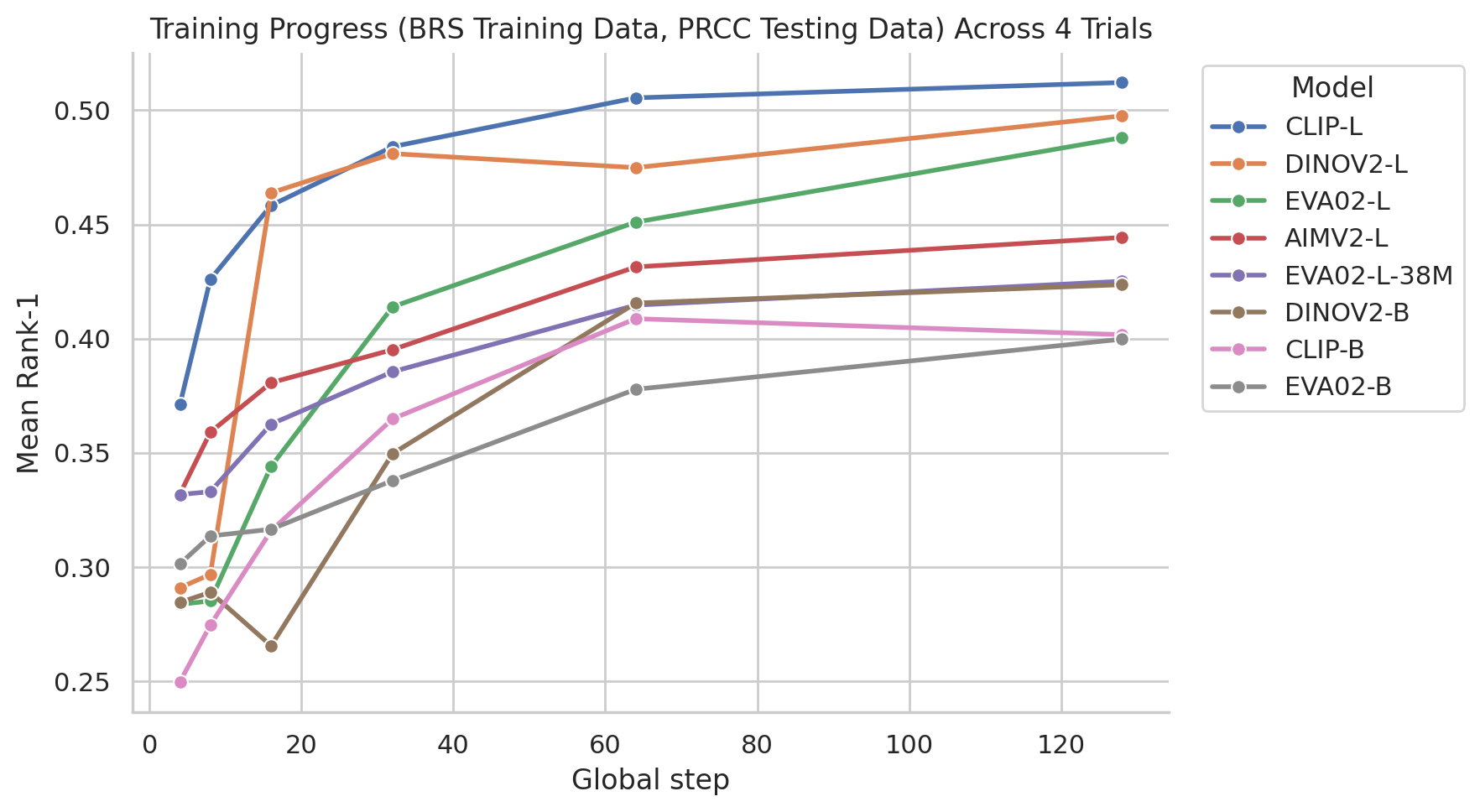}
    \caption{Direct Tuning of foundation models achieves strong, generalizable solutions remarkably quickly.}
    \label{PRCC_efficiency}
\end{figure}

\textbf{Transfer Steps:}
We test the impact of the number of transfer steps required to achieve good re-id solutions for foundation models. Results in Figure \ref{PRCC_efficiency} demonstrate that with the exception of the EVA-02 class of foundation models all fine-tuned foundation models achieve 89+$\%$ of their final rank 1 accuracy after full fine-tuning with only $\approx 5\%$ of the training time and data (full fine-tuning: >2.6k parameter updates, >900k images; abbreviated training: 128 updates, $\approx 40.9$k images). Extending training to 188 steps (60.1k images) improves performance for EVA-02(B) and EVA-02(L), achieving $91.8\%$, and $85.6\%$ of their final rank 1, respectively. The EVA-02(L) variant pre-trained with 38M \cite{Fang_2024} still lags at 188 steps. The decreased sample efficiency may be attributable to the fact that the pre-trained EVA-02 checkpoints include supervised ImageNet training while the other foundations are strictly unsupervised. Regardless, successful adaptation of foundation models to re-id requires relatively small amounts of data and relatively few transfer steps. This further supports the notion that priors inherited from pre-trained foundation weights exist near strong re-id solutions.


\textbf{Weight Decay:}
We test whether use of weight-decay has a strong impact on the solutions reached. We transfer CLIP(L), CLIP(B), and DINOv2(B) using the study-pooled procedure, but set the weight decay to 0. CLIP(L) and CLIP (B) perform substantially worse on PRCC rank-1 and BTS rank-1 while DinoV2 performs slightly better on both metrics (see Table \ref{tab:WD_Results}). Notably, the loss behavior is unchanged (i.e., all three models achieve very low training loss). This suggests that foundation models can achieve more than one low-loss solution of varying quality from a single prior. These results highlight a continued need for moderate regularization when transferring foundations to re-id, even with strong pre-trained priors. 



\begin{table}[h]
\scriptsize
\centering
\begin{tabular}{lcc|cc}
\hline
& \multicolumn{2}{c|}{PRCC Rank 1} & \multicolumn{2}{c}{BTS Fast Rank 1} \\
\cline{2-5}
Model & WD = $1E{-}6$ & WD = 0 & WD = $1E{-}6$ & WD = 0 \\
\hline
DinoV2(B) & .37 & .38 & .36 & .38 \\
CLIP(B)   & .45 & .35 & .35 & .29 \\
CLIP(L)   & .56 & .48 & .44 & .49 \\
\hline
\end{tabular}
\caption{Model Performance With and Without Weight Decay}
\label{tab:WD_Results}
\end{table}

\textbf{Choice of optimizer impacts learned solution:}
We test whether choice of optimizer has a strong impact on the solution reached. We transfer CLIP(L), CLIP(B), and DINOv2(B) using the study pooled procedure but change the optimizer from ADAM to SGD, clipped-SGD, LARS, RMSP, and ADAGRAD. We find it challenging to find high quality solutions with SGD, clipped-SGD, or LARS. We find moderately strong solutions with ADAGRAD, and inconsistent results with RSMP sometimes yielding strong solutions and sometimes yielding poor solutions. Solutions are consistently best with the ADAM optimizer. These results suggest that the geometry of the loss landscape plays a large role in the solution quality. The failure of SGD and LARS suggest an ill-conditioned loss landscape that suffers without per-parameter adaptivity. The success of ADAM over ADAGRAD and RSMP suggests that per-paremeter scaling and momentum are important to learning good re-id solutions from pre-trained foundations, at least under a triplet objective. Notably, RSMP and ADAGRAD still achieve very low training loss re-enforcing that multiple low loss solutions of varying quality can be reached from a single prior.  

\textbf{Choice of loss:}
We test whether choice of loss function has a strong impact on the solution reached. We transfer CLIP(L), CLIP(B), and DINOv2(B) using the Study-Pooled procedure but change the loss to cross-entropy loss. 
Using cross-entropy loss, models achieve low loss solutions; however, when compared to triplet loss, these solutions generalize worse. A triplet loss objective yields rank 1 PRCC scores roughly 25 points better for CLIP(L) ($\approx$.55 vs $\approx$.30), 20 points for CLIP(B)($\approx$.45 vs $\approx$.25), and 4 points for DinoV2(B) ($\approx$.37 vs $\approx$.33). These results suggest that good task-objective alignment is an important feature of transfering foundation models for person identification.   

\textbf{Use of BRS Data:}
We test whether foundation models achieve SOTA solutions on a range of datasets. We transfer the best models from Table \ref{tab:model_comparison} (CLIP(L) and EVA02(L) using the Study-PRCC and Study-Market procedures. The CLIP(L) and EVA02(L) models transferred with Study-PRCC yield .733 and .707 on rank-1 respectively, still achieving SOTA levels. Likewise the CLIP(L) and EVA02(L) models transferred with Study-Market yield .985 and .984 on rank-1 respectively, still achieving SOTA level. These results suggest that foundations offer a strong prior for re-id that can be transferred with a range of datasets.

\section{Conclusion and Future Work}

We have shown that pre-training acts as a strong prior for each kind of re-id problem (short term, clothes change, and long term) and that adapted foundation models are a good point estimate for the Gibb's posterior. Under a triplet loss objective, with small learning rates, weight decay, and an ADAM optimizer, foundation models can be adapted to yield SOTA performance on a range of re-id tasks, even without access to the best data. Going forward, we believe it is critical that future work in re-id be bench-marked directly against fine-tuned foundation models. Without this shift, it becomes very challenging to decouple methodological gains and algorithmic gains, especially as foundation models continue to rise in popularity in re-id pipelines. 

Our work is still limited in that our observations that some pre-trained models provide better priors for re-id tasks is done post-hoc. Future work to establish prior quality before domain adaptation would be valuable. 

We acknowledge that person re-id can have positive impacts (ex. finding missing children) and negative impacts (ex. unwanted surveillance, racial bias). Ultimately, we believe that better understanding of strong re-id models and how they are achieved can lead to more fair, openly testable models. Our pipeline yields strong results with a very simple approach making the study of model bias and its mitigation more accessible in future work.

\newpage
\bibliographystyle{unsrt}
\bibliography{main}

@misc{jager2025expandingbriardatasetcomprehensive,
      title={Expanding on the BRIAR Dataset: A Comprehensive Whole Body Biometric Recognition Resource at Extreme Distances and Real-World Scenarios (Collections 1-4)}, 
      author={Gavin Jager and David Cornett III and Gavin Glenn and Deniz Aykac and Christi Johnson and Robert Zhang and Ryan Shivers and David Bolme and Laura Davies and Scott Dolvin and Nell Barber and Joel Brogan and Nick Burchfield and Carl Dukes and Andrew Duncan and Regina Ferrell and Austin Garrett and Jim Goddard and Jairus Hines and Bart Murphy and Sean Pharris and Brandon Stockwell and Leanne Thompson and Matthew Yohe},
      year={2025},
      eprint={2501.14070},
      archivePrefix={arXiv},
      primaryClass={cs.CV},
      url={https://arxiv.org/abs/2501.14070}, 
}

@incollection{loper2023smpl,
  title={SMPL: A skinned multi-person linear model},
  author={Loper, Matthew and Mahmood, Naureen and Romero, Javier and Pons-Moll, Gerard and Black, Michael J},
  booktitle={Seminal Graphics Papers: Pushing the Boundaries, Volume 2},
  pages={851--866},
  publisher = {Association for Computing Machinery},
  year={2023}
}

@INPROCEEDINGS{7410490,
  author={Zheng, Liang and Shen, Liyue and Tian, Lu and Wang, Shengjin and Wang, Jingdong and Tian, Qi},
  booktitle={2015 IEEE International Conference on Computer Vision (ICCV)}, 
  title={Scalable Person Re-identification: A Benchmark}, 
  year={2015},
  volume={},
  number={},
  pages={1116-1124},
  doi={10.1109/ICCV.2015.133}}

@article{Bissiri_2016,
   title={A General Framework for Updating Belief Distributions},
   volume={78},
   ISSN={1467-9868},
   url={http://dx.doi.org/10.1111/rssb.12158},
   DOI={10.1111/rssb.12158},
   number={5},
   journal={Journal of the Royal Statistical Society Series B: Statistical Methodology},
   publisher={Oxford University Press (OUP)},
   author={Bissiri, P. G. and Holmes, C. C. and Walker, S. G.},
   year={2016},
   month=Feb, pages={1103–1130} }

@article{KWBRC,
  title={Subject Identification up to 1km: Performer Perspective on the IARPA BRIAR Program},
  author={Scott McCloskey and Brandon RichardWebster and Roddy Collins and Anthony Hoogs},
  journal={Proceedings of the National Security Sensor and Data Fusion Committee (NSSDF)},
  volume={},
  pages={},
  year={2023}
}

@article{russakovsky2015imagenet,
  title={Imagenet large scale visual recognition challenge},
  author={Russakovsky, Olga and Deng, Jia and Su, Hao and Krause, Jonathan and Satheesh, Sanjeev and Ma, Sean and Huang, Zhiheng and Karpathy, Andrej and Khosla, Aditya and Bernstein, Michael and others},
  journal={International journal of computer vision},
  volume={115},
  pages={211--252},
  year={2015},
  publisher={Springer}
}

@ARTICLE{9336268,
  author={Ye, Mang and Shen, Jianbing and Lin, Gaojie and Xiang, Tao and Shao, Ling and Hoi, Steven C. H.},
  journal={IEEE Transactions on Pattern Analysis and Machine Intelligence}, 
  title={Deep Learning for Person Re-Identification: A Survey and Outlook}, 
  year={2022},
  volume={44},
  number={6},
  pages={2872-2893},
  doi={10.1109/TPAMI.2021.3054775}}

@article{hahn2016dissecting,
  title={Dissecting the time course of person recognition in natural viewing environments},
  author={Hahn, Carina A and O'Toole, Alice J and Phillips, P Jonathon},
  journal={British Journal of Psychology},
  volume={107},
  number={1},
  pages={117--134},
  year={2016},
  publisher={Wiley Online Library}
}

@inproceedings{huang2025vills,
  title={VILLS: Video-Image Learning to Learn Semantics for Person Re-Identification},
  author={Huang, Siyuan and Prabhakar, Ram and Guo, Yuxiang and Chellappa, Rama and Peng, Cheng},
  booktitle={2025 IEEE/CVF Winter Conference on Applications of Computer Vision (WACV)},
  pages={5969--5979},
  year={2025},
  organization={IEEE}
}

@inproceedings{cornett2023expanding,
  title={Expanding Accurate Person Recognition to New Altitudes and Ranges: The BRIAR Dataset},
  author={Cornett, David and Brogan, Joel and Barber, Nell and Aykac, Deniz and Baird, Seth and Burchfield, Nicholas and Dukes, Carl and Duncan, Andrew and Ferrell, Regina and Goddard, Jim and others},
  booktitle={Proceedings of the IEEE/CVF Winter Conference on Applications of Computer Vision},
  pages={593--602},
  year={2023}
}

@misc{li2023clipreidexploitingvisionlanguagemodel,
      title={CLIP-ReID: Exploiting Vision-Language Model for Image Re-Identification without Concrete Text Labels}, 
      author={Siyuan Li and Li Sun and Qingli Li},
      year={2023},
      eprint={2211.13977},
      archivePrefix={arXiv},
      primaryClass={cs.CV},
      url={https://arxiv.org/abs/2211.13977}, 
}

@misc{han2024clipscgisynthesizedcaptionguidedinversion,
      title={CLIP-SCGI: Synthesized Caption-Guided Inversion for Person Re-Identification}, 
      author={Qianru Han and Xinwei He and Zhi Liu and Sannyuya Liu and Ying Zhang and Jinhai Xiang},
      year={2024},
      eprint={2410.09382},
      archivePrefix={arXiv},
      primaryClass={cs.CV},
      url={https://arxiv.org/abs/2410.09382}, 
}

@misc{zhao2025cilpfgdiexploitingvisionlanguagemodel,
      title={CILP-FGDI: Exploiting Vision-Language Model for Generalizable Person Re-Identification}, 
      author={Huazhong Zhao and Lei Qi and Xin Geng},
      year={2025},
      eprint={2501.16065},
      archivePrefix={arXiv},
      primaryClass={cs.CV},
      url={https://arxiv.org/abs/2501.16065}, 
}

@inproceedings{yu2020cocas,
  title={Cocas: A large-scale clothes changing person dataset for re-identification},
  author={Yu, Shijie and Li, Shihua and Chen, Dapeng and Zhao, Rui and Yan, Junjie and Qiao, Yu},
  booktitle={Proceedings of the IEEE/CVF conference on computer vision and pattern recognition},
  pages={3400--3409},
  year={2020}
}

@misc{dosovitskiy2021imageworth16x16words,
      title={An Image is Worth 16x16 Words: Transformers for Image Recognition at Scale}, 
      author={Alexey Dosovitskiy and Lucas Beyer and Alexander Kolesnikov and Dirk Weissenborn and Xiaohua Zhai and Thomas Unterthiner and Mostafa Dehghani and Matthias Minderer and Georg Heigold and Sylvain Gelly and Jakob Uszkoreit and Neil Houlsby},
      year={2021},
      eprint={2010.11929},
      archivePrefix={arXiv},
      primaryClass={cs.CV},
      url={https://arxiv.org/abs/2010.11929}, 
}

@InProceedings{He_2021_ICCV,
    author    = {He, Shuting and Luo, Hao and Wang, Pichao and Wang, Fan and Li, Hao and Jiang, Wei},
    title     = {TransReID: Transformer-Based Object Re-Identification},
    booktitle = {Proceedings of the IEEE/CVF International Conference on Computer Vision (ICCV)},
    month     = {October},
    year      = {2021},
    pages     = {15013-15022}
}

@misc{sharma2021personreidentificationlocallyaware,
      title={Person Re-Identification with a Locally Aware Transformer}, 
      author={Charu Sharma and Siddhant R. Kapil and David Chapman},
      year={2021},
      eprint={2106.03720},
      archivePrefix={arXiv},
      primaryClass={cs.CV},
      url={https://arxiv.org/abs/2106.03720}, 
}

@misc{liu2021swintransformerhierarchicalvision,
      title={Swin Transformer: Hierarchical Vision Transformer using Shifted Windows}, 
      author={Ze Liu and Yutong Lin and Yue Cao and Han Hu and Yixuan Wei and Zheng Zhang and Stephen Lin and Baining Guo},
      year={2021},
      eprint={2103.14030},
      archivePrefix={arXiv},
      primaryClass={cs.CV},
      url={https://arxiv.org/abs/2103.14030}, 
}

@misc{radford2021learningtransferablevisualmodels,
      title={Learning Transferable Visual Models From Natural Language Supervision}, 
      author={Alec Radford and Jong Wook Kim and Chris Hallacy and Aditya Ramesh and Gabriel Goh and Sandhini Agarwal and Girish Sastry and Amanda Askell and Pamela Mishkin and Jack Clark and Gretchen Krueger and Ilya Sutskever},
      year={2021},
      eprint={2103.00020},
      archivePrefix={arXiv},
      primaryClass={cs.CV},
      url={https://arxiv.org/abs/2103.00020}, 
}

@misc{myers2025unconstrainedbodyrecognitionaltitude,
      title={Unconstrained Body Recognition at Altitude and Range: Comparing Four Approaches}, 
      author={Blake A Myers and Matthew Q Hill and Veda Nandan Gandi and Thomas M Metz and Alice J O'Toole},
      year={2025},
      eprint={2502.07130},
      archivePrefix={arXiv},
      primaryClass={cs.CV},
      url={https://arxiv.org/abs/2502.07130}, 
}

@article{Fang_2024,
   title={EVA-02: A visual representation for neon genesis},
   volume={149},
   ISSN={0262-8856},
   url={http://dx.doi.org/10.1016/j.imavis.2024.105171},
   DOI={10.1016/j.imavis.2024.105171},
   journal={Image and Vision Computing},
   publisher={Elsevier BV},
   author={Fang, Yuxin and Sun, Quan and Wang, Xinggang and Huang, Tiejun and Wang, Xinlong and Cao, Yue},
   year={2024},
   month=sep, pages={105171} }

@misc{su2023roformerenhancedtransformerrotary,
      title={RoFormer: Enhanced Transformer with Rotary Position Embedding}, 
      author={Jianlin Su and Yu Lu and Shengfeng Pan and Ahmed Murtadha and Bo Wen and Yunfeng Liu},
      year={2023},
      eprint={2104.09864},
      archivePrefix={arXiv},
      primaryClass={cs.CL},
      url={https://arxiv.org/abs/2104.09864}, 
}

@misc{metz2025dissectinghumanbodyrepresentations,
      title={Dissecting Human Body Representations in Deep Networks Trained for Person Identification}, 
      author={Thomas M Metz and Matthew Q Hill and Blake Myers and Veda Nandan Gandi and Rahul Chilakapati and Alice J O'Toole},
      year={2025},
      eprint={2502.15934},
      archivePrefix={arXiv},
      primaryClass={cs.CV},
      url={https://arxiv.org/abs/2502.15934}, 
}

@misc{peng2024maskedattributedescriptionembedding,
      title={Masked Attribute Description Embedding for Cloth-Changing Person Re-identification}, 
      author={Chunlei Peng and Boyu Wang and Decheng Liu and Nannan Wang and Ruimin Hu and Xinbo Gao},
      year={2024},
      eprint={2401.05646},
      archivePrefix={arXiv},
      primaryClass={cs.CV},
      url={https://arxiv.org/abs/2401.05646}, 
}

@article{oquab2023dinov2,
  title={Dinov2: Learning robust visual features without supervision},
  author={Oquab, Maxime and Darcet, Timoth{\'e}e and Moutakanni, Th{\'e}o and Vo, Huy and Szafraniec, Marc and Khalidov, Vasil and Fernandez, Pierre and Haziza, Daniel and Massa, Francisco and El-Nouby, Alaaeldin and others},
  journal={arXiv preprint arXiv:2304.07193},
  year={2023}
}

@inproceedings{10.1145/3474085.3475202,
author = {Zhang, Guowen and Zhang, Pingping and Qi, Jinqing and Lu, Huchuan},
title = {HAT: Hierarchical Aggregation Transformers for Person Re-identification},
year = {2021},
isbn = {9781450386517},
publisher = {Association for Computing Machinery},
address = {New York, NY, USA},
url = {https://doi.org/10.1145/3474085.3475202},
doi = {10.1145/3474085.3475202},
booktitle = {Proceedings of the 29th ACM International Conference on Multimedia},
pages = {516–525},
numpages = {10},
location = {Virtual Event, China},
series = {MM '21}
}

@misc{luo2021selfsupervisedpretrainingtransformerbasedperson,
      title={Self-Supervised Pre-Training for Transformer-Based Person Re-Identification}, 
      author={Hao Luo and Pichao Wang and Yi Xu and Feng Ding and Yanxin Zhou and Fan Wang and Hao Li and Rong Jin},
      year={2021},
      eprint={2111.12084},
      archivePrefix={arXiv},
      primaryClass={cs.CV},
      url={https://arxiv.org/abs/2111.12084}, 
}

@misc{li2024clipdrivenclothagnosticfeaturelearning,
      title={CLIP-Driven Cloth-Agnostic Feature Learning for Cloth-Changing Person Re-Identification}, 
      author={Shuang Li and Jiaxu Leng and Guozhang Li and Ji Gan and Haosheng chen and Xinbo Gao},
      year={2024},
      eprint={2406.09198},
      archivePrefix={arXiv},
      primaryClass={cs.CV},
      url={https://arxiv.org/abs/2406.09198}, 
}

@INPROCEEDINGS{10655020,
  author={Liu, Feng and Kim, Minchul and Ren, Zhiyuan and Liu, Xiaoming},
  booktitle={2024 IEEE/CVF Conference on Computer Vision and Pattern Recognition (CVPR)}, 
  title={Distilling CLIP with Dual Guidance for Learning Discriminative Human Body Shape Representation}, 
  year={2024},
  volume={},
  number={},
  pages={256-266},
  doi={10.1109/CVPR52733.2024.00032}}

@misc{ramachandran2017searchingactivationfunctions,
      title={Searching for Activation Functions}, 
      author={Prajit Ramachandran and Barret Zoph and Quoc V. Le},
      year={2017},
      eprint={1710.05941},
      archivePrefix={arXiv},
      primaryClass={cs.NE},
      url={https://arxiv.org/abs/1710.05941}, 
}

@misc{dauphin2017languagemodelinggatedconvolutional,
      title={Language Modeling with Gated Convolutional Networks}, 
      author={Yann N. Dauphin and Angela Fan and Michael Auli and David Grangier},
      year={2017},
      eprint={1612.08083},
      archivePrefix={arXiv},
      primaryClass={cs.CL},
      url={https://arxiv.org/abs/1612.08083}, 
}

@misc{shazeer2020gluvariantsimprovetransformer,
      title={GLU Variants Improve Transformer}, 
      author={Noam Shazeer},
      year={2020},
      eprint={2002.05202},
      archivePrefix={arXiv},
      primaryClass={cs.LG},
      url={https://arxiv.org/abs/2002.05202}, 
}

@misc{hendrycks2023gaussianerrorlinearunits,
      title={Gaussian Error Linear Units (GELUs)}, 
      author={Dan Hendrycks and Kevin Gimpel},
      year={2023},
      eprint={1606.08415},
      archivePrefix={arXiv},
      primaryClass={cs.LG},
      url={https://arxiv.org/abs/1606.08415}, 
}

@misc{wang2022foundationtransformers,
      title={Foundation Transformers}, 
      author={Hongyu Wang and Shuming Ma and Shaohan Huang and Li Dong and Wenhui Wang and Zhiliang Peng and Yu Wu and Payal Bajaj and Saksham Singhal and Alon Benhaim and Barun Patra and Zhun Liu and Vishrav Chaudhary and Xia Song and Furu Wei},
      year={2022},
      eprint={2210.06423},
      archivePrefix={arXiv},
      primaryClass={cs.LG},
      url={https://arxiv.org/abs/2210.06423}, 
}

@InProceedings{pmlr-v9-glorot10a,
  title = 	 {Understanding the difficulty of training deep feedforward neural networks},
  author = 	 {Glorot, Xavier and Bengio, Yoshua},
  booktitle = 	 {Proceedings of the Thirteenth International Conference on Artificial Intelligence and Statistics},
  pages = 	 {249--256},
  year = 	 {2010},
  editor = 	 {Teh, Yee Whye and Titterington, Mike},
  volume = 	 {9},
  series = 	 {Proceedings of Machine Learning Research},
  address = 	 {Chia Laguna Resort, Sardinia, Italy},
  month = 	 {13--15 May},
  publisher =    {PMLR},
  url = 	 {https://proceedings.mlr.press/v9/glorot10a.html}
}

@misc{gu2022clotheschangingpersonreidentificationrgb,
      title={Clothes-Changing Person Re-identification with RGB Modality Only}, 
      author={Xinqian Gu and Hong Chang and Bingpeng Ma and Shutao Bai and Shiguang Shan and Xilin Chen},
      year={2022},
      eprint={2204.06890},
      archivePrefix={arXiv},
      primaryClass={cs.CV},
      url={https://arxiv.org/abs/2204.06890}, 
}

@misc{qian2020longtermclothchangingpersonreidentification,
      title={Long-Term Cloth-Changing Person Re-identification}, 
      author={Xuelin Qian and Wenxuan Wang and Li Zhang and Fangrui Zhu and Yanwei Fu and Tao Xiang and Yu-Gang Jiang and Xiangyang Xue},
      year={2020},
      eprint={2005.12633},
      archivePrefix={arXiv},
      primaryClass={cs.CV},
      url={https://arxiv.org/abs/2005.12633}, 
}

@INPROCEEDINGS{9710001,
  author={Huang, Yan and Wu, Qiang and Xu, JingSong and Zhong, Yi and Zhang, ZhaoXiang},
  booktitle={2021 IEEE/CVF International Conference on Computer Vision (ICCV)}, 
  title={Clothing Status Awareness for Long-Term Person Re-Identification}, 
  year={2021},
  volume={},
  number={},
  pages={11875-11884},
  doi={10.1109/ICCV48922.2021.01168}}

@INPROCEEDINGS{10203842,
  author={Yang, Zhengwei and Lin, Meng and Zhong, Xian and Wu, Yu and Wang, Zheng},
  booktitle={2023 IEEE/CVF Conference on Computer Vision and Pattern Recognition (CVPR)}, 
  title={Good is Bad: Causality Inspired Cloth-debiasing for Cloth-changing Person Re-identification}, 
  year={2023},
  volume={},
  number={},
  pages={1472-1481},
  doi={10.1109/CVPR52729.2023.00148}}

@inproceedings{han2023clothing,
  title={Clothing-change feature augmentation for person re-identification},
  author={Han, Ke and Gong, Shaogang and Huang, Yan and Wang, Liang and Tan, Tieniu},
  booktitle={Proceedings of the IEEE/CVF conference on computer vision and pattern recognition},
  pages={22066--22075},
  year={2023}
}

@article{myers2023recognizing,
      title={Recognizing People by Body Shape Using Deep Networks of Images and Words}, 
      author={Blake A. Myers and Lucas Jaggernauth and Thomas M. Metz and Matthew Q. Hill and Veda Nandan Gandi and Carlos D. Castillo and Alice J. O'Toole},
     journal={Proceedings of the IEEE: International Joint Conference on Biometrics},
     year={2023}
}

@article{huang2024whole,
  title={Whole-Body Detection, Identification and Recognition at Altitude and Range},
  author={Huang, Siyuan and Kathirvel, Ram Prabhakar and Guo, Yuxiang and Lau, Chun Pong and Chellappa, Rama},
  journal={IEEE Transactions on Biometrics, Behavior, and Identity Science},
  year={2024},
  publisher={IEEE}
}

@inproceedings{gu2022clothes,
  author    = {X. Gu and H. Chang and B. Ma and S. Bai and S. Shan and X. Chen},
  title     = {Clothes-Changing Person Re-Identification with RGB Modality Only},
  booktitle = {Proceedings of the IEEE/CVF Conference on Computer Vision and Pattern Recognition (CVPR)},
  year      = {2022},
  pages     = {1060--1069}
}

@INPROCEEDINGS{9959509,
  author={Lee, Kyung Won and Jawade, Bhavin and Mohan, Deen and Setlur, Srirangaraj and Govindaraju, Venu},
  booktitle={2022 18th IEEE International Conference on Advanced Video and Signal Based Surveillance (AVSS)}, 
  title={Attribute De-biased Vision Transformer (AD-ViT) for Long-Term Person Re-identification}, 
  year={2022},
  volume={},
  number={},
  pages={1-8},
  doi={10.1109/AVSS56176.2022.9959509}}

@INPROCEEDINGS{9707584,
  author={Bansal, Vaibhav and Foresti, Gian Luca and Martinel, Niki},
  booktitle={2022 IEEE/CVF Winter Conference on Applications of Computer Vision Workshops (WACVW)}, 
  title={Cloth-Changing Person Re-identification with Self-Attention}, 
  year={2022},
  volume={},
  number={},
  pages={602-610},
  doi={10.1109/WACVW54805.2022.00066}}

@inproceedings{fang2023eva,
  title={Eva: Exploring the limits of masked visual representation learning at scale},
  author={Fang, Yuxin and Wang, Wen and Xie, Binhui and Sun, Quan and Wu, Ledell and Wang, Xinggang and Huang, Tiejun and Wang, Xinlong and Cao, Yue},
  booktitle={Proceedings of the IEEE/CVF conference on computer vision and pattern recognition},
  pages={19358--19369},
  year={2023}
}

@misc{xu2022deepchangelargelongtermperson,
      title={DeepChange: A Large Long-Term Person Re-Identification Benchmark with Clothes Change}, 
      author={Peng Xu and Xiatian Zhu},
      year={2022},
      eprint={2105.14685},
      archivePrefix={arXiv},
      primaryClass={cs.CV},
      url={https://arxiv.org/abs/2105.14685}, 
}

@inproceedings{chen2023beyond,
  title={Beyond Appearance: a Semantic Controllable Self-Supervised Learning Framework for Human-Centric Visual Tasks},
  author={Weihua Chen and Xianzhe Xu and Jian Jia and Hao Luo and Yaohua Wang and Fan Wang and Rong Jin and Xiuyu Sun},
  booktitle={The IEEE/CVF Conference on Computer Vision and Pattern Recognition},
  year={2023},
}

@inproceedings{Mu_2022_BMVC,
author    = {Jingyi Mu and Yong Li and Jun Li and Jian Yang},
title     = {Learning Clothes-irrelevant Cues for Clothes-Changing Person Re-identification},
booktitle = {33rd British Machine Vision Conference 2022, {BMVC} 2022, London, UK, November 21-24, 2022},
publisher = {{BMVA} Press},
year      = {2022},
url       = {https://bmvc2022.mpi-inf.mpg.de/0337.pdf}
}

@ARTICLE{10814721,
  author={Peng, Chunlei and Wang, Boyu and Liu, Decheng and Wang, Nannan and Hu, Ruimin and Gao, Xinbo},
  journal={IEEE Transactions on Multimedia}, 
  title={Masked Attribute Description Embedding for Cloth-Changing Person Re-Identification}, 
  year={2025},
  volume={27},
  number={},
  pages={1475-1485},
  doi={10.1109/TMM.2024.3521730}}

@inproceedings{10.1007/978-3-031-19781-9_12,
author = {Zhu, Kuan and Guo, Haiyun and Yan, Tianyi and Zhu, Yousong and Wang, Jinqiao and Tang, Ming},
title = {PASS: Part-Aware Self-Supervised Pre-Training for Person Re-Identification},
year = {2022},
isbn = {978-3-031-19780-2},
publisher = {Springer-Verlag},
address = {Berlin, Heidelberg},
url = {https://doi.org/10.1007/978-3-031-19781-9_12},
doi = {10.1007/978-3-031-19781-9_12},
booktitle = {Computer Vision – ECCV 2022: 17th European Conference, Tel Aviv, Israel, October 23–27, 2022, Proceedings, Part XIV},
pages = {198–214},
numpages = {17},
location = {Tel Aviv, Israel}
}

@InProceedings{Liang_2025_CVPR,
    author    = {Liang, Xin and Rawat, Yogesh S},
    title     = {DIFFER: Disentangling Identity Features via Semantic Cues for Clothes-Changing Person Re-ID},
    booktitle = {Proceedings of the IEEE/CVF Conference on Computer Vision and Pattern Recognition (CVPR)},
    month     = {June},
    year      = {2025},
    pages     = {13980-13989}
}

@misc{he2015deepresiduallearningimage,
      title={Deep Residual Learning for Image Recognition}, 
      author={Kaiming He and Xiangyu Zhang and Shaoqing Ren and Jian Sun},
      year={2015},
      eprint={1512.03385},
      archivePrefix={arXiv},
      primaryClass={cs.CV},
      url={https://arxiv.org/abs/1512.03385}, 
}

@article{NKUP,
author = {Wang, Kai and Ma, Zhi and Chen, Shiyan and Yang, Jinni and Zhou, Keke and Li, Tao},
title = {A benchmark for clothes variation in person re-identification},
journal = {International Journal of Intelligent Systems},
volume = {35},
number = {12},
pages = {1881-1898},
doi = {https://doi.org/10.1002/int.22276},
url = {https://onlinelibrary.wiley.com/doi/abs/10.1002/int.22276},
eprint = {https://onlinelibrary.wiley.com/doi/pdf/10.1002/int.22276},
year = {2020}
}

@inproceedings{huang2019celebrities,
  author    = {Y. Huang and Q. Wu and J. Xu and Y. Zhong},
  title     = {{Celebrities-reid: A benchmark for clothes variation in long-term person re-identification}},
  booktitle = {2019 International Joint Conference on Neural Networks (IJCNN)},
  year      = {2019},
  publisher = {IEEE},
  pages     = {1--8},
}

@article{shu2021large,
  author    = {X. Shu and X. Wang and X. Zang and S. Zhang and Y. Chen and G. Li and Q. Tian},
  title     = {{Large-scale spatio-temporal person re-identification: Algorithms and benchmark}},
  journal   = {IEEE Transactions on Circuits and Systems for Video Technology},
  volume    = {32},
  number    = {7},
  pages     = {4390--4403},
  year      = {2021},
}

@misc{fini2024multimodalautoregressivepretraininglarge,
      title={Multimodal Autoregressive Pre-training of Large Vision Encoders}, 
      author={Enrico Fini and Mustafa Shukor and Xiujun Li and Philipp Dufter and Michal Klein and David Haldimann and Sai Aitharaju and Victor Guilherme Turrisi da Costa and Louis Béthune and Zhe Gan and Alexander T Toshev and Marcin Eichner and Moin Nabi and Yinfei Yang and Joshua M. Susskind and Alaaeldin El-Nouby},
      year={2024},
      eprint={2411.14402},
      archivePrefix={arXiv},
      primaryClass={cs.CV},
      url={https://arxiv.org/abs/2411.14402}, 
}

\newpage

\appendix
\section{Appendix}

\subsection{Frozen Features on PRCC}
\begin{figure}[H]
\centering
    \includegraphics[scale=0.4]{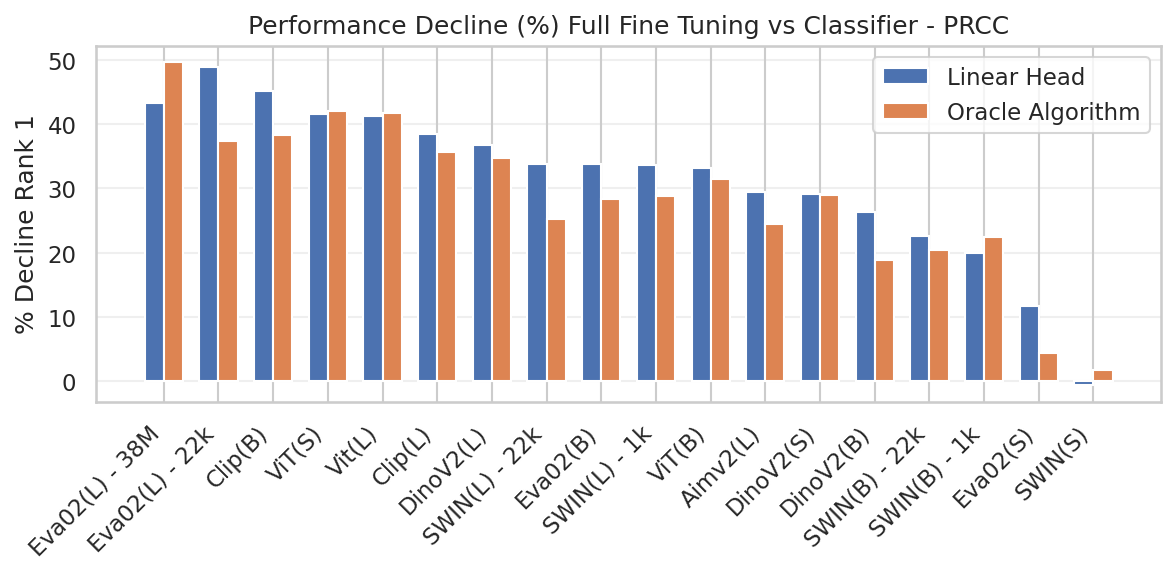}
    \caption{Frozen model features with a trained classier fall short of domain adapted models on PRCC.}
    \label{optim1}
\end{figure}

\subsection{Impact of Optimizer}
\begin{figure}[H]
    \centering
    \includegraphics[scale=0.4]{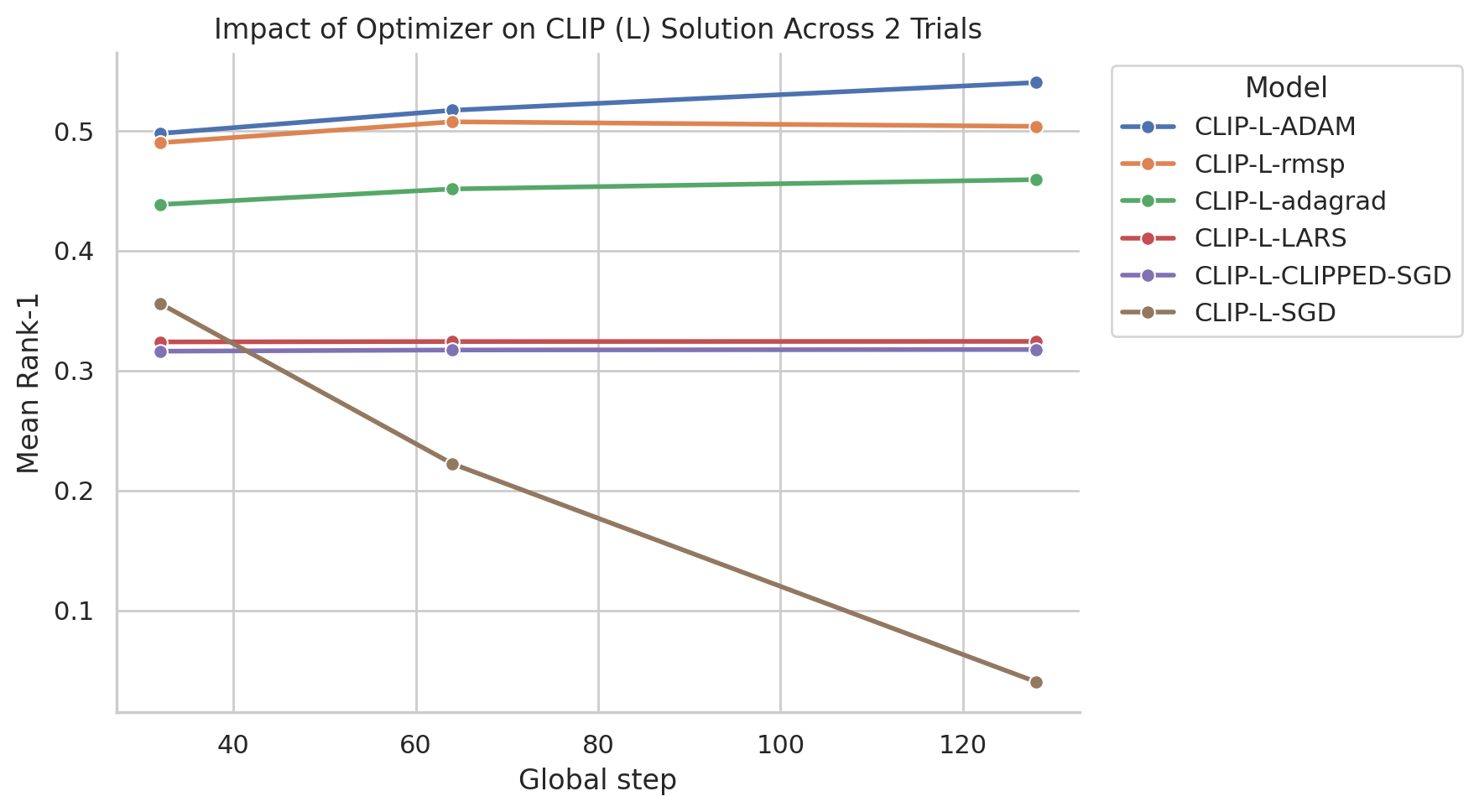}
    \caption{Optimizer has a strong impact on the solution quality achieved with direct Fine Tuning for CLIP(L)}
    \label{optim1}
\end{figure}

\begin{figure}[H]
    \centering
    \includegraphics[scale=0.4]{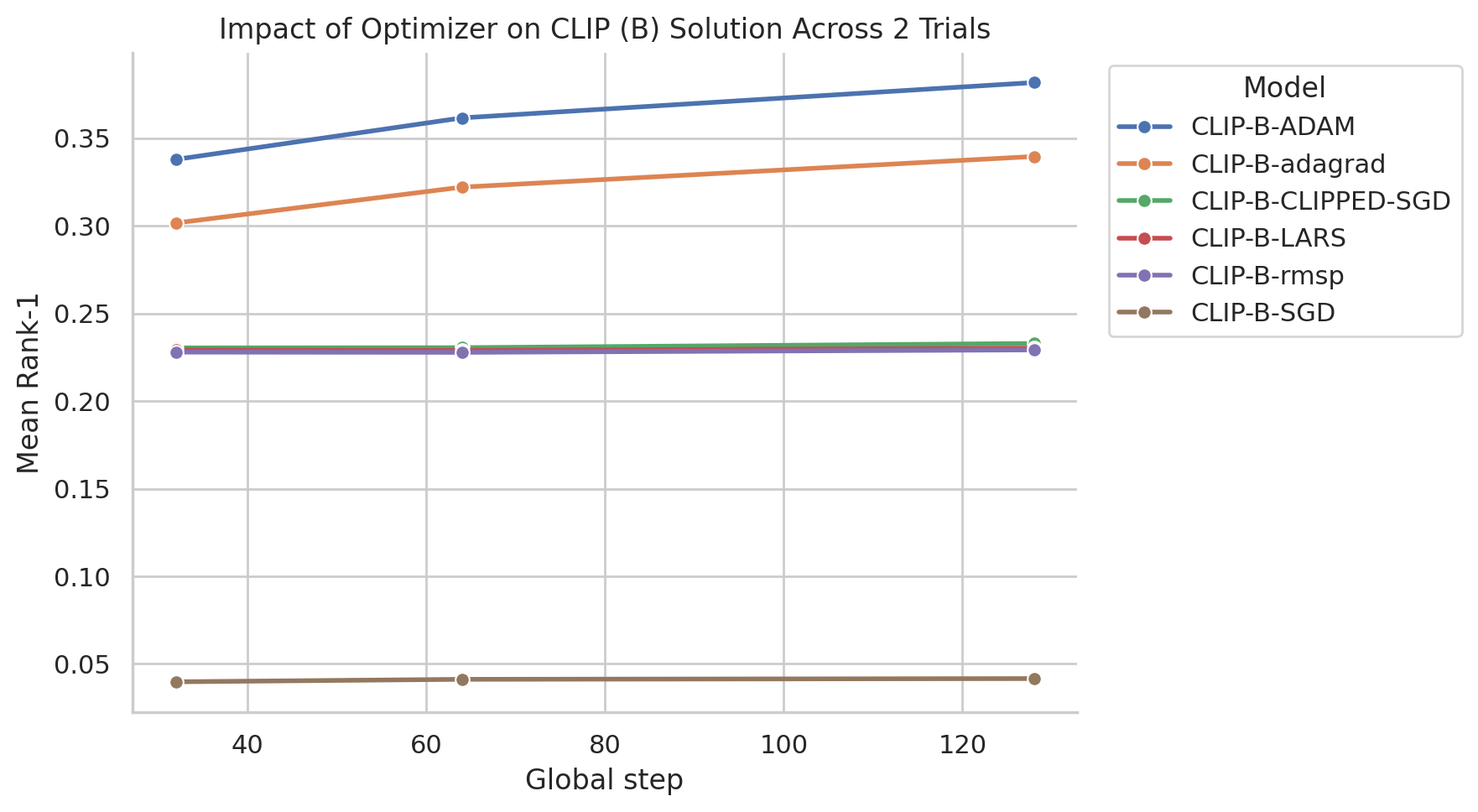}
    \caption{Optimizer has a strong impact on the solution quality achieved with direct Fine Tuning for CLIP(B)}
    \label{2optim}
\end{figure}

\begin{figure}[H]
    \centering
    \includegraphics[scale=0.4]{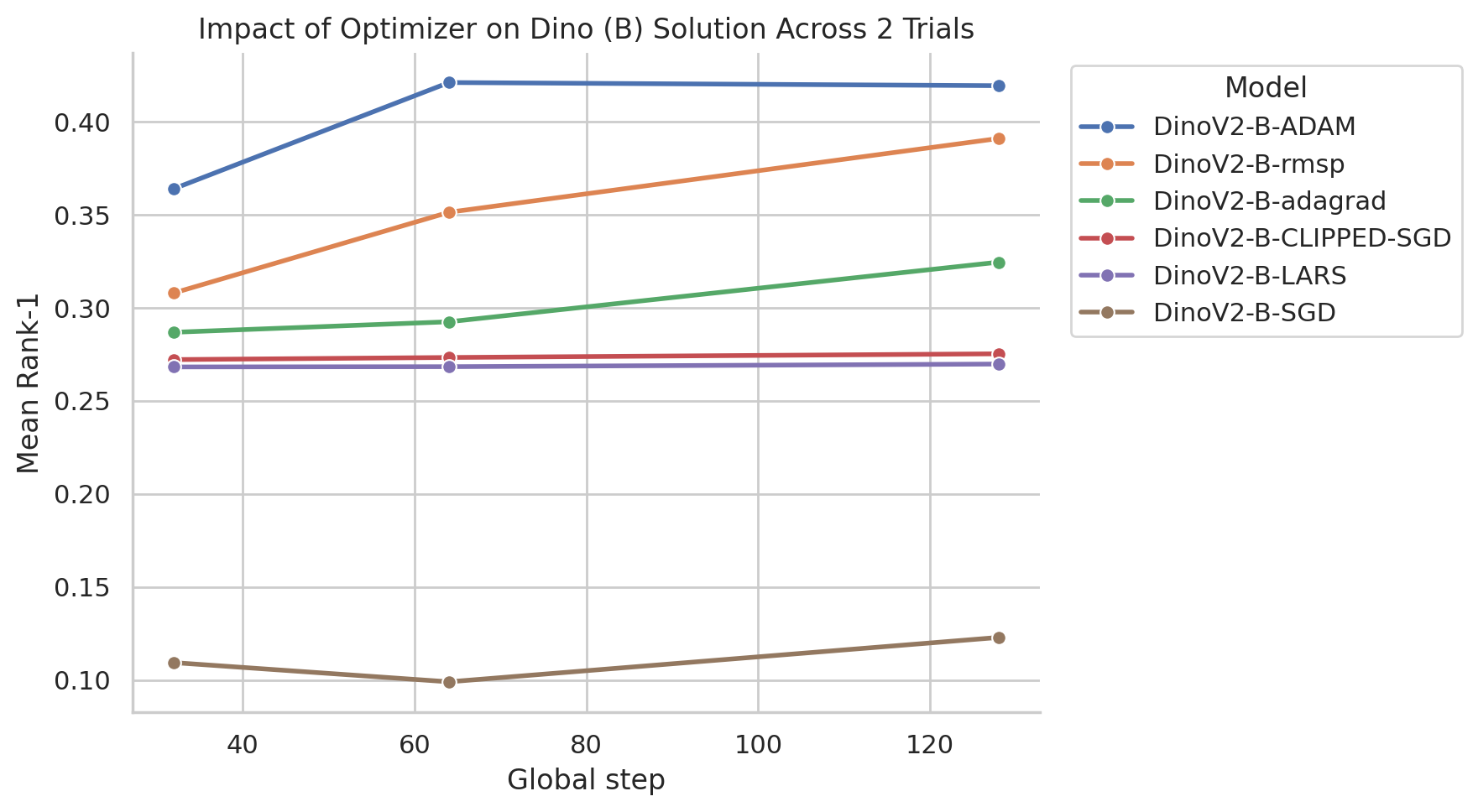}
    \caption{Optimizer has a strong impact on the solution quality achieved with direct Fine Tuning for Dino(B)}
    \label{2optim}
\end{figure}

\subsection{Loss Curves Cross Entropy}
\begin{figure}[H]
    \centering
    \includegraphics[scale=0.4]{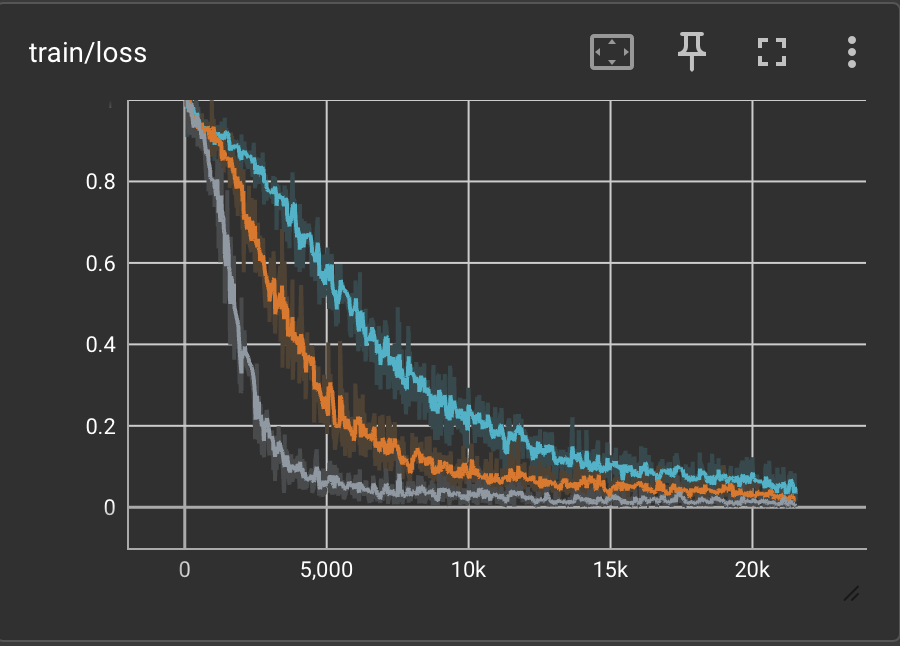}
    \caption{Low loss can be achieved with cross entropy loss with parameters matching the study-pooled protocol but solutions transfer worse. Blue = DinoV2(B), Orange = CLIP(B), Gray = CLIP(L).}
    \label{2optim}
\end{figure}

\subsection{Loss Curves Weight Decay}
\begin{figure}[H]
    \centering
    \includegraphics[scale=0.4]{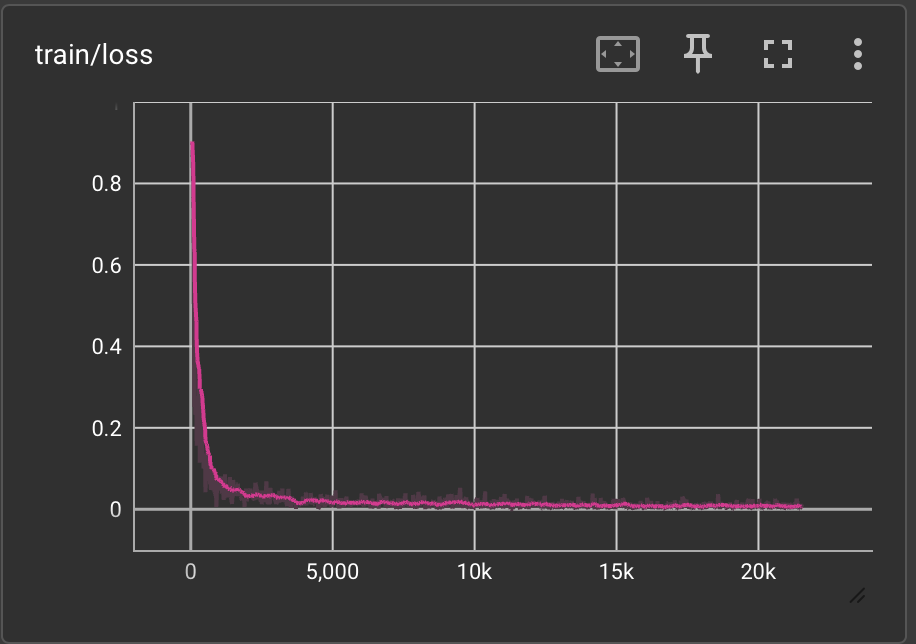}
    \caption{Low loss can be achieved with WD = 0 with study-pooled transfer for DinoV2(B) with little change to solution generalization.}
    \label{2optim}
\end{figure}

\begin{figure}[H]
    \centering
    \includegraphics[scale=0.4]{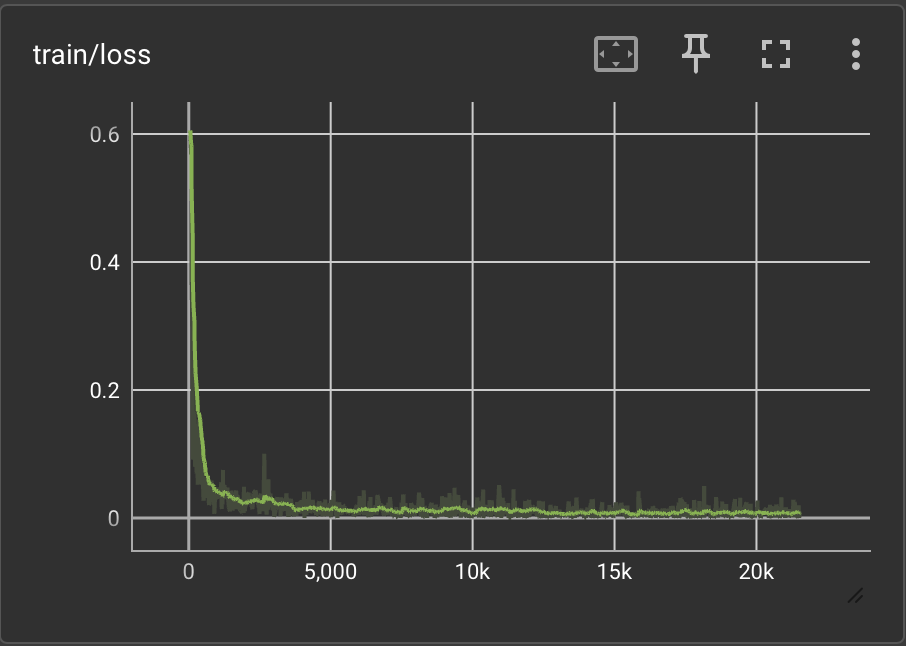}
    \caption{Low loss can be achieved with WD = 0 with study-pooled transfer for CLIP(B) but solutions generalize worse.}
    \label{2optim}
\end{figure}

\begin{figure}[H]
    \centering
    \includegraphics[scale=0.4]{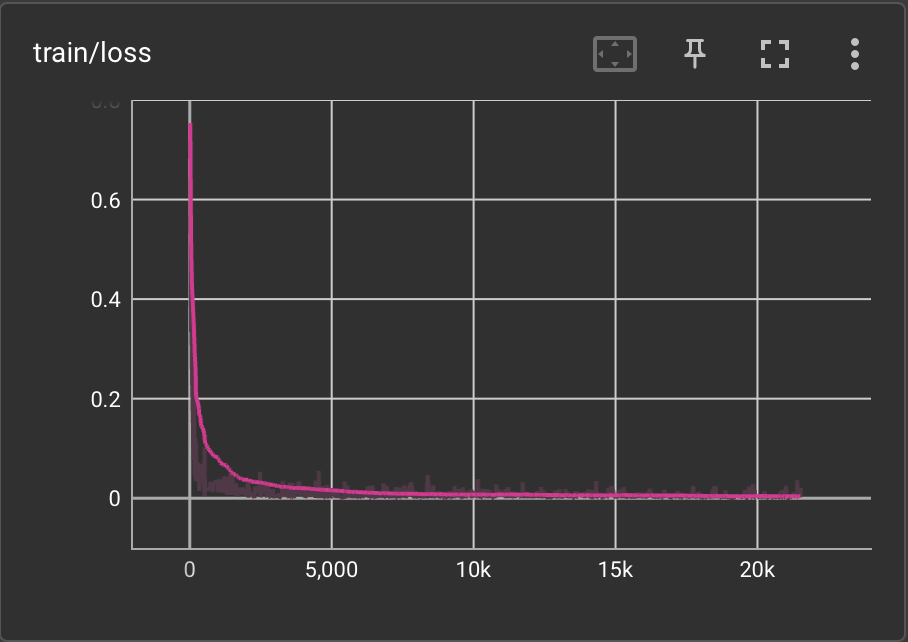}
    \caption{Low loss can be achieved with WD = 0 with study-pooled transfer for CLIP(L) but solutions generalize worse.}
    \label{2optim}
\end{figure}

\end{document}